\documentclass[letterpaper, 10 pt, conference]{ieeeconf}  

\IEEEoverridecommandlockouts                              

\usepackage{algorithm}
\usepackage{algorithmic}
\usepackage{url}
\usepackage{amsmath}
\usepackage{cite}
\usepackage[dvips]{graphicx}
\usepackage{epsfig}
\usepackage{epic,eepic,color}
\usepackage{times}
\usepackage{mathptmx}
\usepackage{multirow}
\usepackage{stkernel}

\usepackage{cases}

\usepackage{times}
\usepackage{multirow}

\definecolor{red}{rgb}{1,0,0}
\definecolor{green}{rgb}{0,1,0}
\definecolor{blue}{rgb}{0,0,1}
\definecolor{violet}{rgb}{1,0,1}
\definecolor{cyan}{cmyk}{1,0,0,0}
\definecolor{magenta}{cmyk}{0,1,0,0}
\definecolor{yellow}{cmyk}{0,0,1,0}

\definecolor{white}{rgb}{1,1,1}

\usepackage{jumoline}

\newcommand{\4}{\bf \color{blue}}
\newcommand{\5}{\color{blue}}

\newcommand{\CO}[1]{}

\CO{

\onecolumn
\textwidth=10cm
}

\newcommand{\CommentOut}[1]{}

\begin{document}

\newcommand{\FIG}[3]{
\begin{minipage}[b]{#1cm}
\begin{center}
\includegraphics[width=#1cm]{#2}
{\scriptsize #3}
\end{center}
\end{minipage}
}

\newcommand{\FIGm}[3]{
\begin{minipage}[b]{#1cm}
\begin{center}
\includegraphics[width=#1cm]{#2}\vspace*{-2mm}\\
{\scriptsize #3}
\end{center}
\end{minipage}
}

\newcommand{\FIGR}[3]{
\begin{minipage}[b]{#1cm}
\begin{center}
\includegraphics[angle=-90,clip,width=#1cm]{#2}\vspace*{1mm}
\\
{\scriptsize #3}
\vspace*{1mm}
\end{center}
\end{minipage}
}

\newcommand{\FIGRpng}[5]{
\begin{minipage}[b]{#1cm}
\begin{center}
\includegraphics[bb=0 0 #4 #5, angle=-90,clip,width=#1cm]{#2}\vspace*{1mm}
\\
{\scriptsize #3}
\vspace*{1mm}
\end{center}
\end{minipage}
}

\newcommand{\FIGpng}[5]{
\begin{minipage}[b]{#1cm}
\begin{center}
\includegraphics[bb=0 0 #4 #5, clip, width=#1cm]{#2}\vspace*{-1mm}\\
{\scriptsize #3}
\vspace*{1mm}
\end{center}
\end{minipage}
}

\newcommand{\FIGtpng}[5]{
\begin{minipage}[t]{#1cm}
\begin{center}
\includegraphics[bb=0 0 #4 #5, clip,width=#1cm]{#2}\vspace*{1mm}
\\
{\scriptsize #3}
\vspace*{1mm}
\end{center}
\end{minipage}
}

\newcommand{\FIGRt}[3]{
\begin{minipage}[t]{#1cm}
\begin{center}
\includegraphics[angle=-90,clip,width=#1cm]{#2}\vspace*{1mm}
\\
{\scriptsize #3}
\vspace*{1mm}
\end{center}
\end{minipage}
}

\newcommand{\FIGRm}[3]{
\begin{minipage}[b]{#1cm}
\begin{center}
\includegraphics[angle=-90,clip,width=#1cm]{#2}\vspace*{0mm}
\\
{\scriptsize #3}
\vspace*{1mm}
\end{center}
\end{minipage}
}

\newcommand{\FIGC}[5]{
\begin{minipage}[b]{#1cm}
\begin{center}
\includegraphics[width=#2cm,height=#3cm]{#4}~$\Longrightarrow$\vspace*{0mm}
\\
{\scriptsize #5}
\vspace*{8mm}
\end{center}
\end{minipage}
}

\newcommand{\FIGf}[3]{
\begin{minipage}[b]{#1cm}
\begin{center}
\fbox{\includegraphics[width=#1cm]{#2}}\vspace*{0.5mm}\\
{\scriptsize #3}
\end{center}
\end{minipage}
}

\title{\LARGE \bf 
Self-localization Using Visual Experience Across Domains
}

\author{Tsukamoto Taisho ~~~~ Tanaka Kanji
\thanks{Our work has been supported in part by 
JSPS KAKENHI 
Grant-in-Aid for Young Scientists (B) 23700229,
and for Scientific Research (C) 26330297 (``The realization of next-generation,
discriminative and succinct SLAM technique: PartSLAM'').}
\thanks{T. Tsukamoto and K. Tanaka are with Graduate School of Engineering, University of Fukui, Japan. {\tt\small tnkknj@u-fukui.ac.jp}}%
\vspace*{-5mm}}

\maketitle
\thispagestyle{empty}
\pagestyle{empty}

\newcommand{\tabF}{
\begin{table*}[t]
\begin{center}
\caption{Comparison against CPD.}\label{tab:F}
  \begin{tabular}{l|rrrr|rrrr|rrrr|rrrr|r} \hline
\multirow{2}{*}{
\begin{tabular}{l}
Query \\ \hline
DB \\
\end{tabular}}
& \multicolumn{4}{|c|}{SP} & \multicolumn{4}{|c|}{SU} & \multicolumn{4}{|c|}{AU} &\multicolumn{4}{|c|}{WI} & Avg.    \\ \cline{2-17}
   & AU & WI & SU & SP &  AU & WI & SP & SU & WI & SP & SU & AU & SP & AU & SU & WI & \\ \hline  \hline 
 
   \ \ \ CS-CD
&13.6&7.4&10.4&5.0&12.5&8.7&9.7&4.4&14.9&11.5&13.6&11.9&10.5&18.9&14.0&7.5&10.9\\ \hline
\multirow{2}{*}{
\begin{tabular}{ll}
CPD \\  
 \\
\end{tabular}}
&44.0&29.4&32.9&35.1&42.5&31.1&33.1&24.7&37.6&40.7&36.1&34.8&37.4&43.2&36.0&39.8&36.2\\ \hline
\end{tabular}
\end{center}
\end{table*}
}

\newcommand{\tabG}{
\begin{table*}[t]
\begin{center}
\caption{Scene retrieval performance in mAP.}\label{tab:G}
\begin{tabular}{l|rrrr|rrrr|rrrr|rrrr|r} \hline
\multirow{2}{*}{
\begin{tabular}{l}
Query \\ \hline
DB \\
\end{tabular}}
& \multicolumn{4}{|c|}{SP} & \multicolumn{4}{|c|}{SU} & \multicolumn{4}{|c|}{AU} & \multicolumn{4}{|c|}{WI} & Avg.    \\ \cline{2-17}
& AU & WI & SU & SP & AU & WI & SP & SU & WI & SP & SU & AU & SP & AU & SU & WI & \\ \hline  \hline 
\ \ \ 
\multirow{4}{*}{
\begin{tabular}{ll}
\hspace*{-5mm} TF-IDF \\  
\hspace*{-5mm} FAB-MAP \\  
\hspace*{-5mm} NBNN-SD \\  
\hspace*{-5mm} CD-SD \\
\end{tabular}}
&0.10&0.08&0.11&0.05&0.05&0.09&0.06&0.09&0.02&0.05&0.05&0.07&0.08&0.05&0.06&0.04&0.07\\ \hline
&0.04&0.11&0.04&0.07&0.06&0.08&0.04&0.12&0.05&0.03&0.07&0.12&0.04&0.03&0.01&0.06&0.06\\ \hline
&0.11&0.20&0.23&0.13&0.06&0.19&0.17&0.28&0.12&0.14&0.18&0.09&0.12&0.08&0.16&0.14&0.15\\ \hline
&0.17&0.28&0.31&0.21&0.13&0.30&0.27&0.38&0.21&0.21&0.22&0.04&0.12&0.07&0.22&0.15&0.21\\ \hline
\end{tabular}
\end{center}
\end{table*}
}

\newcommand{\figA}{
\begin{figure}[t]
\begin{center}
\begin{center}
\FIG{8}{path.eps}{}
\end{center}
\caption{Environments and robot trajectories.}\label{fig:A}
\vspace*{-5mm}
\end{center}
\end{figure}
}

\newcommand{\figB}{
\begin{figure}[t]
\begin{center}
\begin{center}
\FIG{8}{sumbnail.eps}{}
\end{center}
\caption{Retrieval results. From left to right, a query image, the ground-truth database image, and images retrieved by the proposed method and by FAB-MAP.}\label{fig:B}
\end{center}
\end{figure}
}

\newcommand{\figCDd}[1]{
\FIG{7}{#1.jpg.eps}{}
}

\newcommand{\figCD}{
\begin{figure}[t]
\begin{center}
\figCDd{grid1}\vspace*{1mm}\\
\figCDd{grid2}\vspace*{1mm}\\
\figCDd{grid3}\vspace*{1mm}\\
\figCDd{grid4}\vspace*{1mm}\\
\figCDd{grid5}\vspace*{1mm}\\
\figCDd{grid6}\vspace*{1mm}\\
\figCDd{grid7}
\caption{Examples of most contributed subimages.}\label{fig:C}
\end{center}
\end{figure}
}

\newcommand{\figD}{
\begin{figure}[t]
\begin{center}
\begin{center}
\FIG{3}{patch1-1.jpg.eps}{}\vspace*{-4mm}\
\FIG{3}{patch1-2.jpg.eps}{}\vspace*{-4mm}\
\FIG{3}{patch1-3.jpg.eps}{}\vspace*{-4mm}\
\caption{ddd}\label{fig:D}\vspace*{-5mm}
\end{center}
\end{center}
\end{figure}
}

\newcommand{\figG}{
\begin{figure}[t]
\begin{center}
\begin{center}
\FIG{4}{patch4-2.jpg.eps}{}\hspace*{-1mm}%
\FIG{4}{patch4-1.jpg.eps}{}\vspace*{3mm}\\
\FIG{7}{patch4-3.jpg.eps}{}\vspace*{-1mm}\\
\caption{Cross-domain scene descriptor (CD-SD).
 Our objective is to take an
input image (top left) and the keypoints extracted from the image (top right) as input, and create a scene descriptor whose spatial layout is globally similar to that of the keypoints in the input image and locally similar to the visual patterns mined from a cross-domain vocabulary, which is a set of raw visual images collected in different domains (season + route). Shown in the lowermost image of Fig. \ref{fig:G}, is a visualization of our scene descriptor, in which visual patterns mined from the vocabulary are placed at the locations of the corresponding keypoints.}\label{fig:G}
\end{center}
\vspace*{-3mm}
\end{center}
\end{figure}
}

\newcommand{\figJ}{
\begin{figure*}[t]
\begin{center}
\begin{minipage}[b]{8.5cm}
\begin{center}
\FIG{7}{patch5-3.jpg.eps}{}\vspace*{4mm}\\
\FIG{4}{patch5-2.jpg.eps}{}
\FIG{4}{patch3-2.jpg.eps}{}\vspace*{1mm}\\
\FIG{7}{patch3-3.jpg.eps}{}\\
\end{center}
\end{minipage}
\begin{minipage}[b]{8.5cm}
\begin{center}
\FIG{7}{patch2-3.jpg.eps}{}\vspace*{4mm}\\
\FIG{4}{patch2-2.jpg.eps}{}
\FIG{4}{patch1-2.jpg.eps}{}\vspace*{1mm}\\
\FIG{7}{patch1-3.jpg.eps}{}\\
\end{center}
\end{minipage}
\caption{Results for scene parsing. Images in the middle row of this figure are pairs of query images and similar database images. Shown in the upper and lower rows of the figure, is the visualization of our scene parsing for the query and the database images, respectively. The essence of our scene parsing is to explain the SIFT descriptors in an input image by using its NN library descriptor from the library images. To visualize which SIFT descriptor is explained by which library image, we show a synthesized image where a small (size: 20 ~ 20 pixels) patch containing a SIFT descriptor in an input image is replaced by a patch of the same size mined from the library image.}\label{fig:J}
\end{center}
\end{figure*}
}

\newcommand{\figI}{
\begin{figure}[t]
\begin{center}
\begin{center}
\FIG{3}{patch5-1.jpg.eps}{}\vspace*{-4mm}\
\FIG{3}{patch5-2.jpg.eps}{}\vspace*{-4mm}\

\caption{iii}\label{fig:I}
\end{center}
\end{center}
\end{figure}
}

\newcommand{\figK}{
\begin{figure}[t]
\begin{center}
\begin{center}
\FIG{8}{seasons.eps}{}
\caption{Cross-season image datasets, 
acquired 
in autumn (AU:2013/10), 
winter (WI:2013/12), 
spring (SP:2014/4), 
and
summer (SU:2014/7).}\label{fig:K}
\end{center}
\vspace*{-3mm}
\end{center}
\end{figure}
}

\newcommand{\figL}{
\begin{figure}[t]
\begin{center}
\begin{center}
\FIGR{8}{dist0.eps}{}
\caption{Distribution of approximation errors. Horizontal axis: sorted feature ID [\%]. Vertical axis: approximation error in Euclidean distance.
}\label{fig:L}
\end{center}
\end{center}
\end{figure}
}

\newcommand{\figZ}{
\begin{figure}[t]
\begin{center}
\begin{center}
\FIG{7}{freq.eps}{}
\caption{Histograms of visual words used for explaining individual categories (seasons+routes) of input images.
Shown in each row is the histogram for each category of input scene.
``AU,'' ``WI,'' ``SP,'' and ``SU'' are seasons,
while ``1,'' ``2,'' and ``3'' are routes.
}\label{fig:Z}
\end{center}
\vspace*{-3mm}
\end{center}
\end{figure}
}

\newcommand{\tabC}{
\begin{table*}[t]
\begin{center}
\caption{Scene retrieval performance in ANR [\%].}\label{tab:A}
\begin{tabular}{l|rrrr|rrrr|rrrr|rrrr|r} \hline
\multirow{2}{*}{
\begin{tabular}{l}
\hspace*{-2mm} Query \\ \hline
\hspace*{-2mm} DB \\
\end{tabular}}
& \multicolumn{4}{|c|}{SP} & \multicolumn{4}{|c|}{SU} & \multicolumn{4}{|c|}{AU} & \multicolumn{4}{|c|}{WI} & Avg.    \\ \cline{2-17}
& AU & WI & SU & SP & AU & WI & SP & SU & WI & SP & SU & AU & SP & AU & SU & WI & \\ \hline  \hline 
\ \ \ 
\multirow{4}{*}{
\begin{tabular}{ll}
\hspace*{-5mm} TF-IDF \\  
\hspace*{-5mm} FAB-MAP \\  
\hspace*{-5mm} NBNN-SD \\  
\hspace*{-5mm} CD-SD \\ 
\end{tabular}}
&{\4 20.3}&19.2    &26.8    &     16.0&21.8&30.8     &29.9     &    12.5&36.9     &29.9     &25.8     &18.9     &23.0     &{\4 24.0}&29.2     &16.1&23.8\\ \hline
&     25.9&18.5    &34.0    &     16.0&{\4 16.9}&28.4     &24.1     &    13.0&32.1     &24.3     &19.8     &{\4 14.5}&32.2     &30.9     &37.9     &20.2&24.3\\ \hline
&     30.6&12.3    &14.5    &{\5 11.4}&     32.4&21.3     &27.4     &     8.9&24.3     &25.8     &15.9     &23.2     &{\5 17.3}&36.5     &20.5     &{\5 14.5}&21.1\\ \hline
&{\5 23.1}&{\4 6.0}&{\4 9.8}&{\4 8.6} &20.5&{\4 12.2}&{\4 15.1}&{\4 4.5}&{\4 17.5}&{\4 18.2}&{\4 11.4}&24.6     &{\4 15.7}&35.2     &{\4 13.2}&{\4 12.9}&{\4 15.5}\\ \hline
\end{tabular}
\vspace*{-3mm}
\end{center}
\end{table*}
}

\newcommand{\tabE}{
\begin{table*}[t]
\begin{center}
\caption{Results with different types of vocabulary (CD: cross domain, CS: cross season, CR: cross route, FULL: full).}\label{tab:E}
\begin{tabular}{l|rrrr|rrrr|rrrr|rrrr|r} \hline
\multirow{2}{*}{
\begin{tabular}{l}
Query \\ \hline
DB \\
\end{tabular}}
& \multicolumn{4}{|c|}{SP} & \multicolumn{4}{|c|}{SU} & \multicolumn{4}{|c|}{AU} & \multicolumn{4}{|c|}{WI} & Avg.    \\ \cline{2-17}
   & AU & WI & SU & SP & AU & WI & SP & SU & WI & SP & SU & AU & SP & AU & SU & WI & \\ \hline  \hline 
 
 \ \ \ CD
&23.1&6.0&9.6&8.6&20.5&12.2&15.1&4.5&17.5&18.3&11.4&24.6&15.7&35.2&13.2&12.9&15.5\\ \hline
\multirow{4}{*}{
\begin{tabular}{ll}
CS \\  
CR \\  
FULL \\  
 \\ 
\end{tabular}}
&13.1&2.8&6.4&4.1&12.7&6.6&8.9&2.2&12.6&11.3&6.7&14.5&10.4&22.9&7.2&5.8&9.3\\ \hline
&22.1&6.0&9.1&8.1&19.4&10.2&14.2&4.3&17.0&15.9&11.3&24.6&15.7&35.2&13.2&12.9&15.0\\ \hline
&12.5&2.5&5.0&3.7&11.5&6.2&8.2&2.2&12.0&10.8&6.4&14.5&10.4&22.9&7.2&5.8&8.9\\ \hline
\end{tabular}
\end{center}
\end{table*}
}

\begin{abstract}
In this study, we aim to solve the single-view robot self-localization problem by using visual experience across domains. Although the bag-of-words method constitutes a popular approach to single-view localization, it fails badly when it's visual vocabulary is learned and tested in different domains. Further, we are interested in using a cross-domain setting, in which the visual vocabulary is learned in different seasons and routes from the input query/database scenes. Our strategy is to mine a cross-domain visual experience, a library of raw visual images collected in different domains, to discover the relevant visual patterns that effectively explain the input scene, and use them for scene retrieval. In particular, we show that the appearance and the pose of the mined visual patterns of a query scene can be efficiently and discriminatively matched against those of the database scenes by employing image-to-class distance and spatial pyramid matching. Experimental results obtained using a novel cross-domain dataset show that our system achieves promising results despite our visual vocabulary being learned and tested in different domains.
\end{abstract}

\section{Introduction}

In this study, we aim to solve the problem of cross-domain single-view robot self-localization. For solving SLAM and other similar problems in mobile robotics, visual localization is crucial \cite{nnl16,vem3,vem13,nnl2}. While self-localization can be done either by using prior knowledge of the problem domain \cite{vem13} or without them \cite{nnl16}, we deal with applications and scenes where a collection of visual images from different domains, termed cross-domain visual experience, is available as prior knowledge. At the same time, we require the localization algorithm to be extremely fast (to work in a fast robot navigation) and to recognize the place from a single frame \cite{vem13} (i.e., without temporal tracking \cite{burschka2004v} and visual sequence measurements \cite{nnl4}).

One of most popular approaches to address the problem of single-view localization is bag-of-words methods \cite{vem13,nnl2,jegou2009packing}, wherein a collection of local invariant visual features is extracted from an input image, and each feature is translated into a visual word by using a pre-learned library of vector-quantized features. Consequently, an input scene image is described compactly and discriminatively as an unordered collection of visual words (``bag-of-words"). However, as argued by several authors, the bag-of-words method fails badly when learned and tested in different domains; the main reasons include the following: (1) Because the bag-of-words method ignores all information about the spatial layout of the features, it limits the descriptive ability considerably \cite{spm}. 
(2) Essential features are lost during quantization \cite{jegou2009packing}.

\figG

Recently, image-to-class matching techniques have received increasing attention in cross-domain classification tasks \cite{nnl5,nnl6,nnl7,nnl8}. In \cite{nnl5}, 
a novel example-based non-parametric
NBNN classifier was presented. This classifier combines the Naive Bayes assumption with an approximate Parzen estimate. In \cite{nnl6}, the NBNN approach was used for domain adaptation, and it achieved state-of-the-art performance. The NBNN approach is appropriate under the following two conditions: (1) raw visual features are used without vector quantization, and (2) the image-to-class (rather than image-to-image) distances are used for scene comparison. However, in the abovementioned studies, the NBNN approach was used for image classification tasks having pre-defined scene classes. Therefore, the NBNN approach cannot be used directly for those localizations in which there is no explicit scene class; the class has to be learned by the robot itself in an unsupervised manner.

In this study, we address the above issues by mining visual experiences. Our strategy is to mine a cross-domain visual experience (i.e., a library of raw visual images collected in
different domains) to find the relevant visual patterns to effectively explain the input scene. Our approach is motivated by the following facts:
\begin{itemize}
\item
A library of raw images enables a quantization-free approximation of an input scene.
\item
The mined visual features can serve as training data for a given input database scene that characterizes a place.
\item
It is often feasible to explain natural scenes by using visual patterns mined from such an image library \cite{isola2013scene}. 
\end{itemize}
Based on the above idea, 
we propose the approximation of an input local feature 
$f=\langle p, a \rangle$,
which consists of
a keypoint $p$ and an appearance descriptor $a$,
by a set of IDs ---
$l^1$, $\cdots$, $l^K$
of $K$ library features mined from the visual experience,
in the following form:
$f'=\langle p, l^1, \cdots, l^K \rangle$.
This is a compact representation that can be indexed and matched efficiently by an inverted file system.

As the next contribution, we show that the proposed scene descriptor successfully realizes cross-domain localization. In particular, we are interested in a specific cross-domain setting where both the seasons and routes of the visual experience are different from those of the input query/database scenes. Based on the above discussion, we view places (i.e., database images) as independent classes, and for each class, we form a class-specific set of training features, by mining the visual experience to find the relevant library features that effectively explain the input scenes. We show that both the appearance and pose of the mined visual patterns of a query scene can be matched against those of the database scenes by employing image-to-class distance \cite{nnl16} 
and spatial pyramid matching \cite{spm}. 
We conducted experiments using a new ``cross-domain" dataset created from a publicly available image collection 
in our previous ICRA15 paper \cite{icra15a}, 
and found that our system achieved promising results despite our visual library being learned and tested in different domains.

\subsection{Related Work}


In this study, we are interested in cross domain localization rather than scalable localization. 
Many of existing localization frameworks focus on scalable localization that is characterized by a large-size vocabulary.
As an instance, for success of previous BoW methods \cite{vem13} 
(including its variants for soft assignment and multiple assignment per word) in scalable localization, a library of vector quantized visual features is trained and serves as a quantizer. However, this also imposes a limitation, vector quantization errors as aforementioned. In contrast, in our approach, each library feature is directly stored without being approximated nor vector quantized.

On the other extreme, in \cite{vem3}, a simple solution to localization ---using a ``bag-of-raw-features" which matches raw SIFT-like features directly, rather than their vector-quantized representation, is presented. However, this may not be possible in practice when the database size increases because of the high dimensionality of raw SIFT-like descriptors. Our approach can be viewed as exploiting this type of raw feature matching, not for direct matching between query and database images but rather for mining an available visual experience 
to find discriminative visual landmarks. 

The problem of cross domain localization has been attracting increasing attention in recent years \cite{icra2015ws,morita2005view,fremen,sfu_dataset,change_removal,lazy_matching}.
\cite{fremen} is based on the assumption
that some of the mid- to long-term processes that cause the environment changes are (pseudo-)periodic, e.g., seasonal foliage variations, daily illumination cycle and routine human activities.
In \cite{sfu_dataset}, a calibrated, synchronized, and ground truth-aligned dataset of woodland trail navigation in semi-structured and changing outdoor environments is presented. 
The study described in
\cite{change_removal} addressed the problem of change removal and presented a novel approach to learning about appearance change and generalizing the learned change to new locations.
In \cite{lazy_matching},
a lazy-sequences matching algorithm under substantial appearance changes
was presented.
Very recently,
in \cite{linegar2015work},
a localization approach using a map of path memories, ``visual experiences", 
where an experience is a single representation
of the environment under particular conditions,  much like a snapshot, 
is presented.
However,
none of the existing works deal with
the cross domain localization
from a novel perspective of fast single-view localization.

In the literature, NBNN techniques \cite{nnl5,nnl6,nnl7,nnl8} have been mainly studied in the context of image categorization and 
classification with pre-defined scene classes (e.g., place categorization \cite{nnl19}). 
In \cite{nnl5}, the concept of NBNN was introduced by extending 
NN techniques to satisfy conditions (1) and (2) stated above, 
and an improved image classification performance was achieved. 
In \cite{nnl7}, 
the NBNN framework was 
extended to a kernelized version of NBNN. 
In \cite{nnl8}, pooling strategies were 
introduced into the NBNN framework. 
Most relevant to our study is \cite{nnl6}, 
in which the NBNN technique was 
used for domain adaptation and achieved state-of-the-art performance 
in addressing the cross-domain image categorization problem. 
In contast, we cast
our task as a robot self-localization problem
which requires the localization algorithm to be extremely fast,
and in which there is no explicit scene class (i.e., place); 
the class has to be learned by the robot itself, in an incremental manner.

Conceptually, our approach is motivated by our previous work on visual experience mining in ICRA15, IROS14, and VPRiCE15 papers \cite{icra15a,iros14,kanji_vprice}, and differs from all the above works on bag-of-words, direct matching, image-to-class matching and spatial pyramid matching. This study is also different from our previous work on ``cross-season" localization in \cite{icra15a} 
where the vocabulary is learned and tested in mutually overlapping routes.
To the best of our knowledge, these issues have not been explored in existing work.

\section{Approach}\label{sec:approach}

\subsection{Library of Visual Experience}

As the first novelty of our approach,
a library of raw image data for local feature vocabulary, 
is used as a prior without vector quantization. 
The library images are 
not required to be associated with spatial information such that the viewpoint and orientation are known. 
%
For example, they
can be visual experiences obtained by the robot itself in a previous navigation, 
images shared by other colleague robots, or 
a publicly available image data resource on the web, 
such as Google StreetView.

\figK

Our ``cross-domain" library consists of view images collected using a handheld monocular camera in a university campus over four different seasons and three different routes (Fig. \ref{fig:K}).
The routes start at three different locations inside the university campus,--- some going through the main central path and others going through the pedestrian walkway along the campus wall, as shown in Fig. \ref{fig:A}.
We considered a typical scenario that deals with view images 
taken relatively far apart (e.g., 1m) from each other.
Severe occlusion occurred in all the scenes, 
which were occupied by 
people and vehicles as dynamic entities.
The datasets also have significant viewpoint change due to their handheld nature.
As can be seen a single place looks quite different 
depending on the geometric conditions (e.g., viewpoints, fresh snow covers) and photometric conditions,
making our localization task
a challenging one. 

\subsection{Scene Descriptor}

Similar to bag-of-words approaches
the proposed scene descriptor
represents an input image
by unordered collections of local features;
However,
we do not rely on a 
library of vector quantized
visual features  (i.e., visual words)
and domain-specific library learning;
instead, we
use a library of $V$ raw visual features: 
\begin{equation}
z[1], \cdots, z[V].
\end{equation}
Our feature library
is obtained 
by computing SIFT local features 
for all images in the image library.

Our scene interpretation step begins by
extracting $N$ SIFT features from the query (or database) scene image.
For each extracted feature $x$,
we search its $K$ nearest neighbor library features 
$S^{K-NN}=$$\{ i_k | i_k \in [1,V] \}_{k=1}^K$ (NN features) 
whose distances from $x$ are shortest over the library,
and describe the search result 
using a $V$-dimensional vector $f^{query}$ (or $f^{database}$)
whose $i$-th element is a truncated similarity:
\begin{equation}
f^{query}[i] = 
\begin{cases}
\max( D_0^2 - || x-z[i] ||^2 , 0 ) & \mbox{(if $i\in S^{K-NN}$)} \\
0 & \mbox{(otherwise)}. \label{eqn:1}
\end{cases}
\end{equation}
\begin{equation}
f^{database}[i] = 
\begin{cases}
1 & \mbox{(if $i\in S^{K'-NN}$)} \\
0 & \mbox{(otherwise)}. \label{eqn:1}
\end{cases}
\end{equation}
The resulted descriptor is a size $N$ set of $V$-dim vectors 
$\{$$f_1$, $\cdots$, $f_N$$\}$
which we term nearest neighbor descriptor.
Since each $V$-dim vector is a sparse vector with only 
$K$ (or $K'$) non-zero elements, our descriptor can be efficiently stored and compared using an inverted file system. In addition, we 
need to store only IDs of non-zero elements for mapped images to compact the database,
and we use both IDs and similarity values for query images.
We fix the parameters $K$ and $D_0$ throughout the paper:
$D_0=200$, 
$K=10$ for query feature, 
and $K'=3$ for database feature. 

For scene comparison, 
we employ the image-to-class (rather than image-to-image) distance with the NBNN formulation, which has proven to be effective for cross domain scene recognition in \cite{nnl6}. 
For this purpose, we view places (i.e., database images) as independent classes, and for each class, 
we prepare a class specific set of training features. 
The training set is obtained by a data mining approach, in which the feature library is mined for discovering NN library features that effectively explain each feature in the database image of interest, and the training set is represented by the mined NN features for each class. 
A comparison between a pair of a query NN descriptor $f^{query}$ and a database NN descriptor $f^{database}$ is based on a similarity function in the form:
\begin{equation}
I(f^{query}, f^{database}) = \max_{i=1}^V \Big( f^{query}[i] \Big) \Big( f^{database}[i] \Big).
\end{equation}
Then,
we solve the scene matching problem
as a search problem in the form:
\begin{equation}
c^* = \arg \max_c \sum_{i=1}^N 
\left(
\max_{j=1}^N I(f^{query}_i, f^{database}_j) 
\right)
\end{equation}
where 
$c$ is a candidate class (i.e., place)
and $x_i$ is the query feature.

\subsection{Spatial Pyramid Matching}

We adopt spatial pyramid matching 
by placing a sequence of increasingly coarser grids over the image region and taking a weighted sum of the number of matches that occur at each level of resolution. At any fixed resolution, two features are compared in terms of a given similarity measure $I$. 
$K$ is the pyramid match kernel, defined as:
\begin{equation}
K(X,Y)=\frac{1}{2^L} I^0 + \sum_{l=1}^L \frac{1}{2^{L-l+1}} I^l \label{eqn:pmk}
\end{equation}
In the original implementation of spatial pyramid matching in \cite{spm}, 
the similarity measure $I$ aims to measure similarity between a pair of bag-of-words vectors in terms of histogram intersection. Note that the number of matches at level $l$ also includes all the matches found at the finer level $l+1$. Therefore, the number of new matches found at level $l$ is given by $I^l-I^{l+1}$ for $l=0$, $\cdots$, $L$. In addition, matches found at coarser level are penalized by using a weight $1/2^l$ which is inversely proportional to the cell width at level $l$. 
In this study, we replace this similarity function $I^l$ that measures similarity in terms of the NBNN based similarity, so that we can incorpolate the robust NBNN distance within the spatial pyramid matching framework, to obtain image level similarity. Note that the similarity function $K$ reduces to a standard NBNN similarity function when $L=0$. The resulted scene descriptor is a sparse long $16V$-dim vector formed by concatenating the NN descriptors at the finest level.

\section{Experiments}

\figA

\tabC
\tabE
\tabF
\tabG

\subsection{Settings}

We collected a new ``cross-domain" dataset 
on three different paths \#1-\#3
in a university campus environment 
as shown in Fig. \ref{fig:A}.
We went each path for three times,
collected three independent collections of images and use each for query, library and database image collection.
The datasets were collected across all the four seasons over a period of one year, as shown in Fig. \ref{fig:K}. In addition, we collected an independent set of 3,537 images in the autumn season on routes different from those above and added it to all the databases considered here, as a set of additional distructer images.

In this study, 
we implemented several comparing methods,
FAB-MAP, TF-IDF, CPD, and an NBNN descriptor w/o spatial pyramid (NBNN-SD) 
and the proposed cross-domain scene-descriptor with spatial pyramid (CD-SD), for performance evaluation.
FAB-MAP is 
appearance based localization framework
based on a bag-of-words scene models
and learning of co-occurance of visual words.
We used the code provided by the authors in \cite{vem13}.
TF-IDF is a standard method for bag-of-words image retrieval.
We follow a setting in \cite{jegou2009packing}
and set the visual dictionary size 10,000.
CPD is the technique we have proposed in previous ICRA15 paper,
where a small collection of 
8 visual phrases 
that effecively explain an input scene is discovered
by a common pattern discovery 
between the input and the visual experience.
We used the same code as in \cite{icra15a}.
NBNN-SD is different from the proposed method only in that 
it does not consider spatial layout of the scene
and this is realized by setting the parameter $L=0$.
Finally, 
CD-SD is the proposed method,
described in previous section, 
Section \ref{sec:approach}.
We set the parameter $L=2$ as a default.

\figL

Fig. \ref{fig:L} shows an investigation of approximation errors of the proposed approach. 
Recalling that our approach approximates each SIFT descriptor in a query image 
by an NN library feature mined from the library of visual experience,
the approximation error is represented by
distance between an input SIFT descriptor 
and the NN library descriptor.
In this study, 
we investigate 
the approximation error induced by cross domain library,
and compare it 
with that induced by non cross domain library.
To this end,
we compared 
distance 
from each query feature 
to the nearest neighbor library feature.

\figB

\figZ

\subsection{Examples of Image Retrieval}

Fig. \ref{fig:B} shows
five examples of image retrieval,
from left to right, 
the input query image, the ground truth image, 
and the database images 
top-ranked 
by the proposed CD-SD method
and 
by the FAB-MAP. 
One can see that 
each method returns database images
that are similar to query image.
However,
the FAB-MAP method tends to fail
when there are 
confusing images in the database
whose appearance is partially similar with the query image
but with different spatial layout.
In constrast,
the proposed method
was successful in these
examples
by using both 
appearance and spatial layout of features
as a discriminative cue.

\subsection{Performance Results}\label{sec:performance}

Table \ref{tab:A} shows performance results.
We evaluated the proposed 
CD-SD method 
and other comparing methods,
TF-IDF, FAB-MAP, and NBNN-SD.
We found that the CPD method was very slow to be tested on a database of this size, and therefore, a comparison with CPD using a relatively small database was reported in Section \ref{sec:cpd}.

Series of independent 
$12 \times 100$ retrievals are conducted 
for each of the 12 combinations of routes and seasons.
The retrieval performance was measured in terms 
of the percentage (\%) averaged normalized rank (ANR)---a ranking-based retrieval performance measure, for which a smaller value indicates better performance. 
The ranking based performance measure
is more suitable for localization task than precision/recall
measure that has been often used in classical image retrieval
tasks which aim at finding as many relevant database images
as possible for a given query image. 
This measure is motivated
by the fact that for the sake of localization, it is
sufficient to retrieve just one relevant image. 

To evaluate ANR, we evaluated the rank assigned to the ground-truth
relevant image for each of the 100 independent retrievals;
we normalized the rank with respect to the database size
and computed the average over the 100 retrievals. For the
ground truth, an image $i$ that is most similar to the query
image is manually selected from the database images and its
neighbors $[i-10; i+10]$ are defined as ground truth images.
Different database is prepared for different query 
and contains only one random ground-truth database image.
Note that our database images contain spatially
dense viewpoints, which makes it difficult for a localizer
to distinguish between them.
From Table \ref{tab:A},
one can see that our approach outperformed the BoW method in most of the retrievals considered here.

Table \ref{tab:G} shows mAP performance.
We observe that 
mAP value tends to be a low value,
as each dataset contains only one ground-truth database image
in our case.
One can see that 
the proposed CD-SD method again
outperforms the other methods considered here.

\subsection{Comparison against CPD}\label{sec:cpd}

We conducted an independent experiment to compare the performance between the proposed CD-SD method and CPD method, due to the reason described in Section \ref{sec:performance}. Table \ref{tab:F} shows the performance results. One can see that the proposed method clearly outperforms the CPD method. A main reason is that the CPD method uses a region-level feature, ``visual phrase", to represent the visual patterns mined from the visual experience, and it often fails to explain an input scene when it looks very dissimilar from library scenes as shown in this experiment.


\subsection{Comparison against Different Vocabularies}\label{sec:voc}

Table \ref{tab:E} 
reports a result of comparing performance results
between cross-domain (CD) vocabulary focused in this paper
and three different vocabularies, cross-season (CS), cross-route (CR), and full (FULL) vocabularies.
CS or CR is a union of vocabularies from different seasons or routes, and thus, is an easier setting than CD. FULL is the easiest setting, in which a union of all the 12 vocabularies is employed. The performance results suggest that localization by using the CR vocabulary is not as bad as localization by using the CD vocabulary. While localization by using the CR vocabulary can be considered easier than localization by using the CD vocabulary, it is challenging compared to localization by using the CS or FULL vocabulary.

\figCD

\figJ

\subsection{Frequency of library images}

Fig. \ref{fig:Z}
summarizes
the frequency of individual vocabularies
being used for explaining
individual scene categories.
In this study, we categorized library images into 12 disjoint classes (i.e., 4 seasons ~ 3 routes), and investigated the frequency of library images from each class used for explaining query images. We summarize the result for each of the 12 disjoint classes and show it in the figure. We can see that the frequency of the most and the least frequently used vocabularies differ by almost three times, and that every vocabulary is used for every case considered in the current experiments.

\subsection{Spatial Pyramid Matching}

We here demonstrate the effect of spatial pyramid matching 
compared to conventional image-to-image matching.
The key idea of spatial pyramid matching
is to match 
sub-images 
at each location
at each level 
between query and database images.
To demonstrate the effect,
we visualize which 
partitions are similar between
query and database images
for several examples of pairs of query and database images.
To this end,
we evaluated 
the sub-image level similarity 
between query and database images
for each of the 
$\sum_{l=0}^24^l=21$ 
pairs of sub-images,
and 
selected top 5 similar sub-image pairs,
i.e.,
5 sub-image pairs
that most contributed to the image-level similarity. 
Fig. \ref{fig:C}
shows the top 5 similar sub-image pairs by overlaying bounding boxes on the original input image. We can see that the largest sub-image that corresponds to the entire image region was always selected as one of the most contributed subimages for all the cases and that salient
subimages of various sizes and locations were selected.

\subsection{NN Descriptor}

Fig. \ref{fig:J} demonstrates scene parsing for 
two pairs of query and database images.
The essence of our scene parsing is to explain the SIFT descriptors in an input image by using its NN library descriptor from the library images. To visualize which SIFT descriptor is explained by which library image, we show a synthesized image where a small (size: 20 $\times$ 20 pixels) patch containing a SIFT descriptor in an input image is replaced by a patch of the same size mined from the library image.
Fig. \ref{fig:J}
shows a visualization result.
It can be seen that 
structure and nature parts
of a scene
are well explained by visual patterns 
mined from very different seasons and places.

\section{Conclusions}


In this study, we aimed to solve the single-view robot self-localization problem by using a cross-domain vocabulary. Further, we are interested in using a cross-domain setting, in which the visual vocabulary is learned in different seasons and routes from the input query/database scenes. Our strategy is to mine a cross-domain visual experience, a library of raw visual images collected in different domains, to discover the relevant visual patterns that effectively explain the input scene, and use them for scene retrieval. In particular, we showed that the appearance and the pose of the mined visual patterns of a query scene can be efficiently and discriminatively matched against those of the database scenes by employing image-to-class distance and spatial pyramid matching. Experimental results obtained using a novel cross-domain dataset showed that our system achieves promising results despite our visual vocabulary being learned and tested in different domains.

\bibliographystyle{IEEEtran}
\bibliography{spm}

\end{document}